\documentclass{article}

\usepackage[preprint]{neurips_2025}

\usepackage[utf8]{inputenc} 
\usepackage[T1]{fontenc}    
\usepackage{hyperref}       
\usepackage{url}            
\usepackage{booktabs}       
\usepackage{amsfonts}       
\usepackage{nicefrac}       
\usepackage{microtype}      
\usepackage{xcolor}         
\usepackage{geometry}

\usepackage{times}
\usepackage{latexsym}

\usepackage{amssymb}

\usepackage[T1]{fontenc}

\usepackage[utf8]{inputenc}

\usepackage{microtype}

\usepackage{inconsolata}

\usepackage{amsmath}
\usepackage[pdftex]{graphicx}
\usepackage{booktabs}
\usepackage{blindtext}
\usepackage{lipsum}
\usepackage{mwe}
\usepackage{adjustbox}
\usepackage{graphicx} 
\usepackage{array}

\let\cite\undefined

\newcommand{\llamaoneb}{\texttt{Llama-3.2-1B}}
\newcommand{\llamathreeb}{\texttt{Llama-3.2-3B}}
\newcommand{\llamaeightb}{\texttt{Llama-3.1-8B}}
\newcommand{\llamaeightbinstruct}{\texttt{Llama-3.1-8B-Instruct}}
\newcommand{\mistralsevenb}{\texttt{Mistral-7B-v0.1}}
\newcommand{\qwensevenb}{\texttt{Qwen2.5-7B}}
\newcommand{\qwenfortheenb}{\texttt{Qwen3-14B}}
\newcommand{\qwenfortheenbinstruct}{\texttt{Qwen3-14B-Instruct}}

\newcommand{\idktunedmistral}{\texttt{IDK}-\texttt{tuned}-\texttt{Mistral-7B-v0.1}}

\title{Pretrained LLMs Learn Multiple Types of Uncertainty}

\author{Roi Cohen \\
         HPI / University of Potsdam\\
        \texttt{Roi.Cohen@hpi.de} \\\And
  Omri Fahn \\
  Tel Aviv University \\
  \texttt{omrifahn@mail.tau.ac.il} \\\And
  Gerard de Melo \\
  HPI / University of Potsdam\\
  \texttt{Gerard.DeMelo@hpi.de}}

\begin{document}

\maketitle

\begin{abstract}
  Large Language Models are known to capture real-world knowledge, allowing them to excel in many downstream tasks. Despite recent advances, these models are still prone to what are commonly known as hallucinations, causing them to emit unwanted and factually incorrect text. In this work, we study how well LLMs capture uncertainty, without explicitly being trained for that. We show that, if considering uncertainty as a linear concept in the model's latent space, it might indeed be captured, even after only pretraining. 
  We further show that, though unintuitive, LLMs appear to capture several  different types of uncertainty, each of which can be useful to predict the correctness for a specific task or benchmark. Furthermore, we provide in-depth results such as demonstrating a correlation between our correction prediction and the model's ability to abstain from misinformation using words, and the lack of impact of model scaling for capturing uncertainty. Finally, we claim that unifying the uncertainty types as a single one using instruction-tuning or [IDK]-token tuning is helpful for the model in terms of correctness prediction. 
\end{abstract}

\section{Introduction}

Large Language Models (LLMs) are trained on vast corpora of text data
\citep{brown2020language, raffel2020exploring, chowdhery2023palm, touvron2023llama, le2023bloom, jiang2023mistral} 
enabling them to comprehend and generate human language. These training datasets encompass a wide range of written human knowledge, including books, news articles, Wikipedia, and scientific publications. Through this extensive pretraining, LLMs retain significant portions of the information they are exposed to, effectively embedding real-world knowledge within their parameters and functioning as knowledge repositories \citep{petroni2019language, roberts2020much, cohen-etal-2023-crawling, LLMsKGs2023}.
This capability allows LLMs to be leveraged in tasks that depend on such knowledge, such as closed-book question answering \citep{brown2020language, roberts2020much} and information retrieval \citep{tay2022transformer}.

Despite their widespread adoption, LLMs are widely known to suffer from `hallucinations'—a predisposition towards producing outputs that are false or misleading—which significantly undermines their accuracy and trustworthiness \citep{10.1145/3571730, ManduchiEtAl2024GenAI}. Hallucinations may manifest in various forms, including factually incorrect statements \citep{maynez-etal-2020-faithfulness, devaraj-etal-2022-evaluating, tam-etal-2023-evaluating}, internal inconsistencies \citep{elazar-etal-2021-measuring, mundler2023self}, contradictions \citep{cohen2024evaluating}, or statements lacking clear sources or attribution \citep{bohnet2022attributed, rashkin2023measuring, yue-etal-2023-automatic}.

Uncertainty, however, is a concept that LLMs are not generally known to capture \citep{yin2023large, kapoor2024large}. At the very least, they are generally not explicitly trained on it. This lack of competency regarding uncertainty, however, often results in misinformation generation, which can be harmful and misleading \citep{maynez-etal-2020-faithfulness, devaraj-etal-2022-evaluating, tam-etal-2023-evaluating}, as LLMs have a hard time expressing a lack of knowledge both verbally and through their output distribution. 

\begin{figure}
\setlength{\belowcaptionskip}{-10pt}
    \centering
    \includegraphics[width=0.84\textwidth]{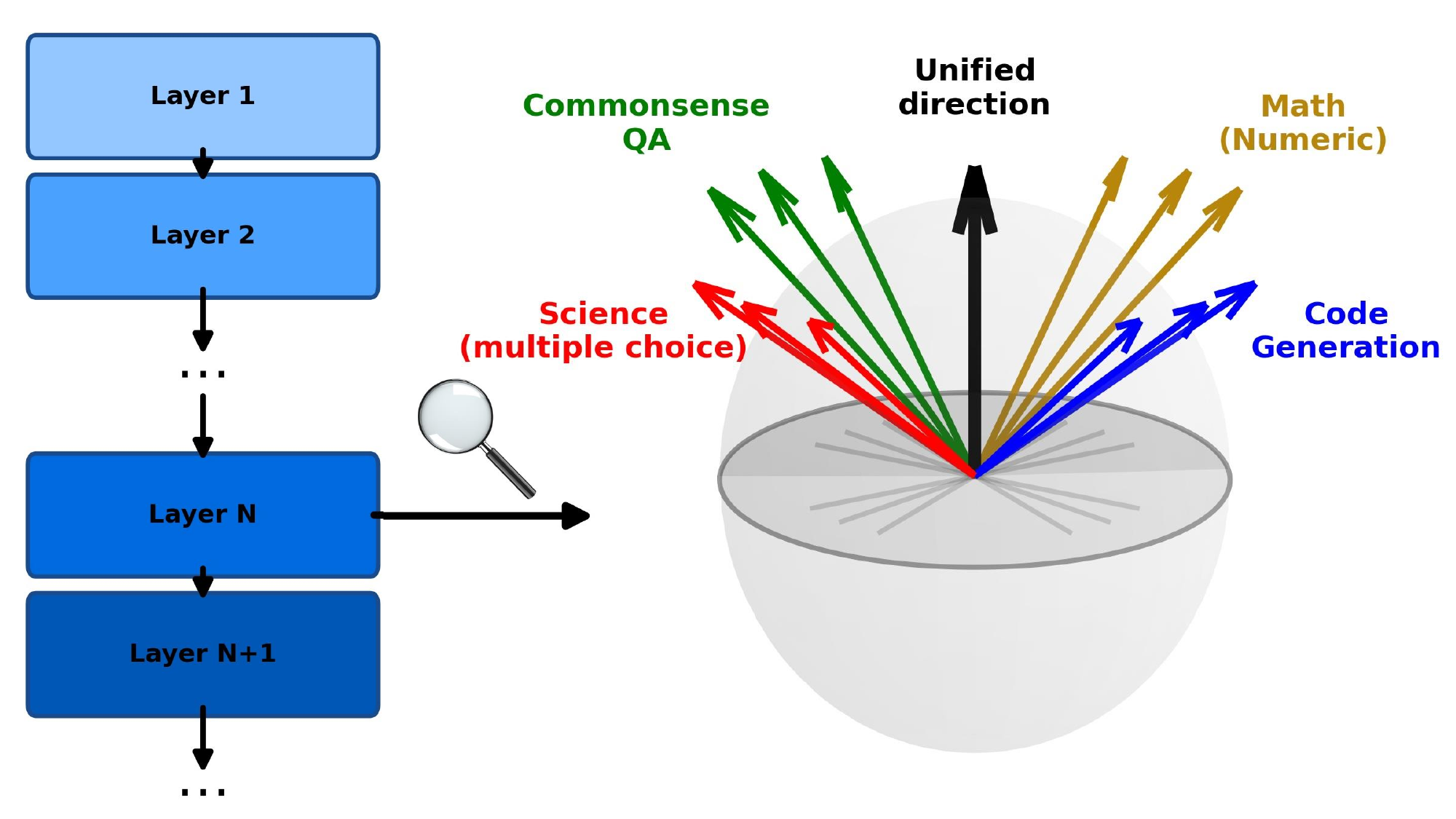}
    \caption{Illustration of identifying multiple data-specific uncertainty linear vectors when investigating the hidden space at the end of each transformer layer.}
    \label{figure:intro}
\end{figure}

Some more advanced methods such as instruction-tuning \citep{ouyang2022training, zhang2023instruction} during post-training and [IDK]-tuning \citep{cohen2024don} during pretraining aim, inter alia, to align LLMs to more efficiently express their uncertainty and refrain from misinformation generation. While instruction tuning more generally aligns LLMs with human intent by fine-tuning them on task-specific instructions and corresponding outputs, the model is often also encouraged to refrain from answering questions when the specific answer is not known to it.

In this work, we propose an analysis mechanism, which we use in order to study the uncertainty captured by a diverse range of models. First, we propose a technique to search for linear vectors in the LLMs' latent space that are associated with uncertainty. We then suggest using these vectors as a form of correctness prediction for the LLM's own generation. By establishing this regime, we can evaluate how well these vectors can stand as misinformation predictors and approximate their uncertainty expression quality. 

Using our proposed mechanism, we demonstrate that LLMs indeed internalize a notion of uncertainty during pretraining, which can be extracted using linear probes from their latent representations. Specifically, we show that it is possible to identify linear uncertainty vectors—directions in the model's hidden space—that correlate with generation correctness across multiple models and datasets, despite forgoing any additional training of model weights. This suggests that uncertainty is a learnable and linearly separable concept within LLMs' latent spaces.

Interestingly, the study reveals that LLMs do not learn a single, unified representation of uncertainty. Instead, they are found to encode multiple distinct uncertainty vectors, each associated with different datasets or types of knowledge. These vectors often exhibit low cosine similarity, indicating near linear independence. However, some generalization exists: For instance, uncertainty representations derived from multiple mathematics benchmarks can transfer across related datasets. These insights may enable new hallucination mitigation techniques, since inconsistencies between learned uncertainty representations may contribute to unreliable or incorrect outputs.
Moreover, we conduct an in-depth analysis in terms of transformer layers, model sizes, and different training techniques. We find that intermediate transformer layers are typically the most informative for extracting uncertainty vectors, consistently yielding the highest accuracy in correctness prediction across datasets. In addition, the model size alone does not appear to enhance uncertainty representation, as smaller models often perform on par with or even surpass larger counterparts in this task. 

More notably, instruction-tuning and [IDK]-tuning significantly boost a model’s ability to capture uncertainty. Instruction-tuned variants of Llama and Qwen models outperform their base versions, and their optimal uncertainty representations also emerge in earlier layers. Similarly, [IDK]-tuning not only improves the overall correctness prediction accuracy but also aligns early layers more effectively with uncertainty signals, as evinced by higher precision in early-stage classifiers. These results suggest that targeted training strategies can enhance the internal encoding of uncertainty more effectively than scaling model size alone.

To conclude, our contributions are: (1) We introduce an analytical framework for probing how LLMs encode uncertainty, (2) we conduct thorough experiments across models, layers, and datasets, and show that uncertainty is not only a learnable and linearly separable concept but also represented in multiple, distinct forms within a single model, (3) we further analyze how factors such as model depth, size, and training methods affect uncertainty representation, revealing that intermediate layers are most informative, and that scaling the model size does not guarantee better uncertainty encoding, and (4) we show that instruction-tuning and [IDK]-tuning significantly improve uncertainty capturing, offering practical strategies for enhancing model reliability and reducing hallucinations.

\section{Identifying Uncertainty Predictors}
\label{sec:setup}

In this work, we assume that uncertainty is a concept represented in an LLM's latent space in each of the layers. Specifically, let $\mathbf{h}_i(x)$ be the hidden state produced by the end of the $i$-th layer of the model, given input $x$. Then, for each of these hidden states, we search for a specific linear vector $\mathbf{u}_i$ such that the classifier defined as $C(x,i) = \mathbf{u}_i^\intercal \mathbf{h}_i(x) + b_i$ can reach an accuracy level of predicting the correctness of the model's next token generation that is  statistically significantly better than random accuracy. Intuitively, this search seeks to identify a linear concept that represents the uncertainty of the model regarding its own generations. 

\subsection{Linear Uncertainty Search}
\label{sec:linear_uncertainty_search}
Let $M$ be a specific LLM and let $\mathcal{D}$ be a specific dataset of questions and answers $\mathcal{D} = \{(q_j, a_j)\}_{j=0}^{n}$. In order to find a certain $\mathbf{u}_i$ for a certain model's layer $i$, we train a straightforward linear classifier for the sake of predicting the correctness of the model's answer to a specific question. Specifically, let $\mathcal{D}_\textsc{Train} = \{(q_j, a_j)\}_{j=0}^{m < n}$ be a training set derived from $\mathcal{D}$. For each question $q_j$ in the dataset, we first let the model predict its own answer. If the model's prediction is correct compared to $a_j$, then we label $q_j$ as positive. In contrast, if the model's prediction is incorrect compared to $a_j$, then we label $q_j$ as negative. Formally, assuming $M(q_j)$ is the model's output given the input $q_j$, we then define its label $L(q_j)$ as:
\begin{equation} \label{eq:labeling}
L(q_j) = \begin{cases} 
1 & \text{if } M(q_j) = a_j \\
 0 & \text{otherwise.} 
\end{cases}
\end{equation}
We thus define our new training set as $\hat{\mathcal{D}}_\textsc{Train} = \{(q_j, L(q_j)\}_{j=0}^{m}$. 
We now can train a classifier at the end of each layer in $M$'s architecture. The input of the classifier is the produced hidden state by the end of the specific layer. As mentioned before, the purpose of this classifier is to predict the correctness of the upcoming prediction of $M$ itself. If we employ a linear classifier, this would correspond to a linear direction in this layer's latent space, which we will refer to as the \emph{uncertainty direction} corresponding to dataset $\mathcal{D}$ (as this direction has been found while training the classifier to predict the correctness of the model on this specific dataset). We thus denote it as $\mathbf{u}_i(\mathcal{D})$. We denote the corresponding learned bias term as $b_i$.

\subsection{Uncertainty Vector as a Predictor}
\label{sec:linear_uncertainty_eval}
We later can evaluate the quality of $\mathbf{u}_i(\mathcal{D})$ by testing its ability to predict the correctness of the model's generation given unseen data as input. Particularly, in this work, we will use test sets derived from our question answering datasets, which we use in order to train our classifier during the linear uncertainty search process (see Section~\ref{sec:linear_uncertainty_search}). Technically, given a textual input $x$ to the model $M$, recall that $\mathbf{h}_i(x)$ is the hidden state vector produced by the end of the $i$-th layer of $M$ during the inference call $M(x)$. We thus, as mentioned before, will use $\mathbf{u}_i(\mathcal{D})$ as a linear classifier in order to predict the correctness of the model's generation -- namely the token that the model $M$ assigns the highest probability as a next-token completion for $x$. More formally, let $C_{\mathbf{u}_i(\mathcal{D})}$ be the classifier induced by $\mathbf{u}_i(\mathcal{D})$ and let $C_{\mathbf{u}_i(\mathcal{D})}(x)$ be the predicted correctness of $x$ while applying $\mathbf{u}_i(\mathcal{D})$. Then:

\begin{equation} \label{regularization_only}
C_{\mathbf{u}_i(\mathcal{D})}(x) = \begin{cases} 
\textsc{Incorrect} & \text{if } [\mathbf{u}_i(\mathcal{D})]^\intercal \mathbf{h}_i(x) + b_i > 0 \\
 \textsc{Correct} & \text{if } [\mathbf{u}_i(\mathcal{D})]^\intercal \mathbf{h}_i(x) + b_i \leq 0 
\end{cases}
\end{equation}

Given the correctness prediction of the uncertainty vector, we can evaluate its correctness in case we have the ground-truth token. We thus can also derive general accuracy, precision, recall, etc.

\begin{table*}[t]
\centering
\setlength\tabcolsep{0.0001pt}
\small
\resizebox{\textwidth}{!}{\begin{tabular}{lccccccccccccccccccc}
\toprule
\textbf{Model} & |ARC-Easy & |ASDiv-A & |CommonsenseQA & |GSM8K & |GranolaEntityQuestions & |HumanEval-X & |MBPP & |NaturalQuestions & |OpenBookQA & |PopQA & |Qampari & |ROMQA & |SVAMP & |StrategyQA & |TriviaQA & |TruthfulQA & |~ \\ \midrule
        \llamaoneb{} & 0.535 & 0.670 & 0.625 & 0.444 & 0.789 & 0.708 & 0.769 & 0.600 & 0.534 & 0.857 & 0.634 & 0.750 & 0.729 & 0.689 & 0.716 & 0.737 & ~ \\ 
        \llamathreeb{} & 0.710 & 0.648 & 0.598 & 0.688 & 0.790 & 0.732 & 0.641 & 0.675 & 0.590 & 0.793 & 0.734 & 0.583 & 0.750 & 0.608 & 0.742 & 0.600 & ~ \\ 
        \llamaeightb{} & 0.657 & 0.667 & 0.649 & 0.577 & 0.763 & 0.692 & 0.722 & 0.590 & 0.644 & 0.757 & 0.630 & 0.763 & 0.711 & 0.684 & 0.757 & 0.722 & ~ \\ \midrule
        \llamaeightbinstruct{} & 0.652 & 0.885 & 0.667 & 0.737 & 0.705 & 0.781 & 0.728 & 0.655 & 0.694 & 0.768 & 0.679 & 0.750 & 0.767 & 0.639 & 0.776 & 0.719 & ~ \\ \midrule
        \mistralsevenb{} & 0.657 & 0.691 & 0.709 & 0.550 & 0.782 & 0.707 & 0.707 & 0.630 & 0.597 & 0.747 & 0.727 & 0.750 & 0.687 & 0.643 & 0.760 & 0.673 & ~ \\ \midrule
        \idktunedmistral{} & 0.600 & 0.750 & 0.571 & 0.688 & 0.758 & 0.545 & 0.688 & 0.673 & 0.611 & 0.829 & 0.789 & 0.667 & 0.628 & 0.547 & 0.693 & 0.725 & ~ \\ \midrule
        \qwensevenb{} & 0.750 & 0.800 & 0.718 & 0.682 & 0.704 & 0.578 & 0.648 & 0.750 & 0.678 & 0.817 & 0.697 & 0.615 & 0.696 & 0.698 & 0.717 & 0.678 & ~ \\ 
        \qwenfortheenb & 0.727 & 0.786 & 0.655 & 0.878 & 0.738 & 0.800 & 0.694 & 0.651 & 0.743 & 0.833 & 0.630 & 0.596 & 0.789 & 0.561 & 0.782 & 0.699 & ~ \\ \midrule
        \qwenfortheenbinstruct{} & 0.800 & 0.750 & 0.638 & 0.702 & 0.770 & 0.688 & 0.625 & 0.674 & 0.619 & 0.771 & 0.861 & 0.655 & 0.767 & 0.711 & 0.756 & 0.726 & ~ \\ 
        
\bottomrule
\end{tabular}}

\caption{Correctness prediction accuracy of our induced classifiers derived across all datasets}
\label{table:acc_all_models_all_datasets_best_layer}
\end{table*}

\section{Experimental Setup}
\label{sec:expermental_setup}

To evaluate our uncertainty identification framework, we consider a series of experiments, for which we first introduce the experimental setup. 

\paragraph{Foundation Models.} In order to reach general conclusions that are not specific to any particular LLM, in this work we study three different families of models -- The Llama family of models \citep{touvron2023llama, dubey2024llama}, Mistral \citep{Jiang2023Mistral7}, and Qwen \citep{Bai2023QwenTR, yang2024qwen2}. Specifically, for Llama we study \llamaoneb{}, \llamathreeb{}, and \llamaeightb{}, for Mistral, we study \mistralsevenb{}, and finally for Qwen, we study \qwensevenb{} and \qwenfortheenb{}. 

\paragraph{Advanced Models.} For evaluating the effects of different types of training on the linear uncertainty encodings, we exploit three particular additional models in our experiments. To capture the instruction-tuning \citep{ouyang2022training, zhang2023instruction} effect we use \llamaeightbinstruct{} and \qwenfortheenbinstruct{}, which both were post-trained in instruction-tuning fashion. Furthermore, we follow \citep{cohen2024don} and use the \idktunedmistral{} model in our experiments to evaluate the effect of [IDK]-tuning -- a method that essentially adds a new uncertainty token to the model's vocabulary and teaches the model to use it during pretraining by adapting its loss to consider the new token.






\paragraph{Datasets and Benchmarks.}
\label{sec:datasets}
We utilize 16 QA datasets and benchmarks in both our linear uncertainty search
(Section~\ref{sec:linear_uncertainty_search}) and the induced classifier evaluation
(Section~\ref{sec:linear_uncertainty_eval}).  We group them into six thematic categories:

\begin{itemize}
  \item \textbf{Commonsense QA}: 
        \textit{CommonsenseQA} \citep{talmor-etal-2019-commonsenseqa}, 
        \textit{StrategyQA} \citep{geva-etal-2021-aristotle}.
        These include questions that assess the model’s ability to apply everyday reasoning and background knowledge to answer questions beyond surface-level facts.
  \item \textbf{Fact-Lookup and Adversarial QA}: 
        \textit{GranolaEntityQuestions} \citep{yona-etal-2024-narrowing}, 
        \textit{Natural Questions} \citep{kwiatkowski-etal-2019-natural}, 
        \textit{PopQA} \citep{mallen2022not}, 
        \textit{TriviaQA} \citep{joshi2017triviaqa}, 
        \textit{TruthfulQA} \citep{Lin2021TruthfulQAMH}.
        These consist of questions that test the model's factual recall and resilience to misleading or adversarial question phrasing.
  \item \textbf{List-Output QA}: 
        \textit{QAMPARI} \citep{amouyal-etal-2023-qampari}, 
        \textit{RoMQA} \citep{Zhong2022RoMQAAB}.
        Both evaluate whether models can produce comprehensive sets of correct answers, challenging their ability to recall multiple relevant facts simultaneously
  \item \textbf{Science QA (K–12)}: 
        \textit{ARC-Easy} \citep{clark2018think}, 
        \textit{OpenBookQA} \citep{mihaylov2018can}.
        These focus on elementary school and high-school level science, requiring models to combine factual knowledge with basic reasoning.
  \item \textbf{Math Word Problems}: 
        \textit{GSM8K} \citep{cobbe2021training}, 
        \textit{ASDiv-A} \citep{miao-etal-2020-diverse}, 
        \textit{SVAMP} \citep{patel-etal-2021-nlp}.
        These include queries that test models on arithmetic and algebraic reasoning through natural language mathematical problems.
  \item \textbf{Code Generation}: 
        \textit{HumanEval-X} \citep{zheng2023codegeex}, 
        \textit{MBPP} \citep{austin2021program}.
        We use these datasets to evaluate the ability of models to generate correct and functional software code given natural language programming prompts.
\end{itemize}

Notably, for each of these, we create a fixed train split which will be used to derive our uncertainty vectors, and a test split which will be used to evaluate their performance. 


\paragraph{Linear Uncertainty Search Details.}

For every model~$M$, transformer layer~$i$, and evaluation dataset~$\mathcal{D}$, we fit a logistic-regression probe on the hidden states $\mathbf{h}_i(x)$ and obtain a single weight vector,
\[
  \mathbf{u}_i(\mathcal{D}),
\]
which serves as the \emph{linear uncertainty direction} for that (layer,\,dataset) pair.

To obtain a dataset-agnostic baseline, we also train an additional probe on the \textbf{concatenation of \emph{all} datasets}.  The resulting vector is denoted as
\[
  \mathbf{u}_i(\mathcal{D}_\textsc{Unified}).
\]

\paragraph{Evaluation.}
We evaluate the ability of our identified uncertainty linear vectors to predict the correctness of the model's generation. For this, we consider the following metrics: (i) \textbf{Accuracy} -- the ratio of correct predictions by the classifier that is induced by the uncertainty linear vector, (ii) \textbf{Precision} -- the ratio of actually wrong completions by the model among those that the induced classifier predicted to be wrong.

\begin{figure}
    \centering
    \begin{minipage}{0.45\textwidth}
        \centering
        \includegraphics[width=1\textwidth]{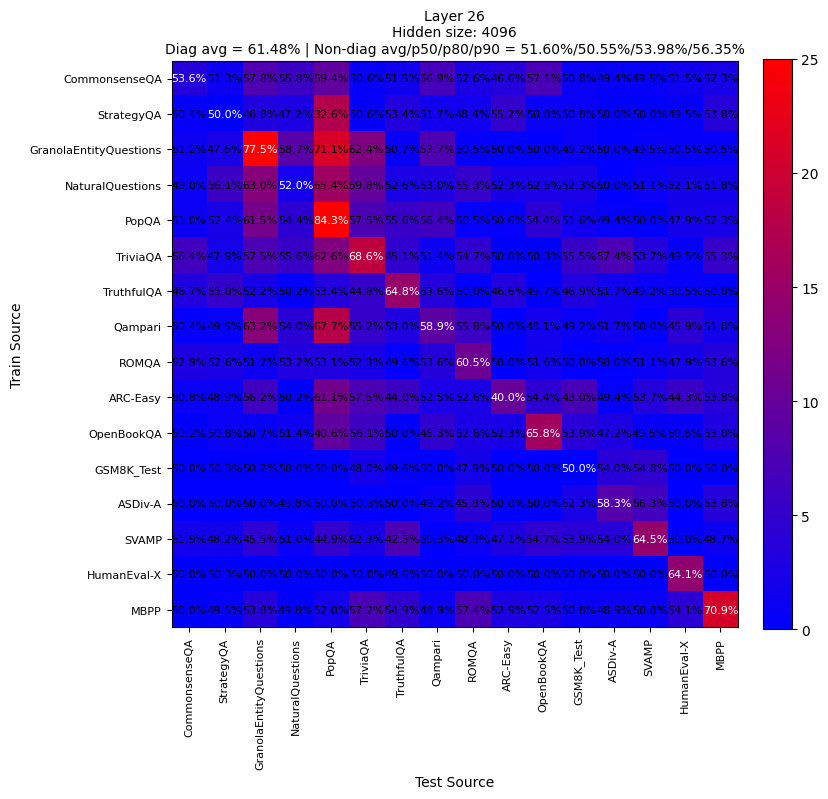} %
        \caption{Correctness prediction accuracy results of the classifier induced by $u_{26}(y-axis-dataset)$, using \llamaeightb{}, while testing on the test set of the x-axis dataset.}
    \label{figure:cross_evaluation_llama8_26}
    \end{minipage}\hfill
    \begin{minipage}{0.45\textwidth}
        \centering
        \includegraphics[width=1\textwidth]{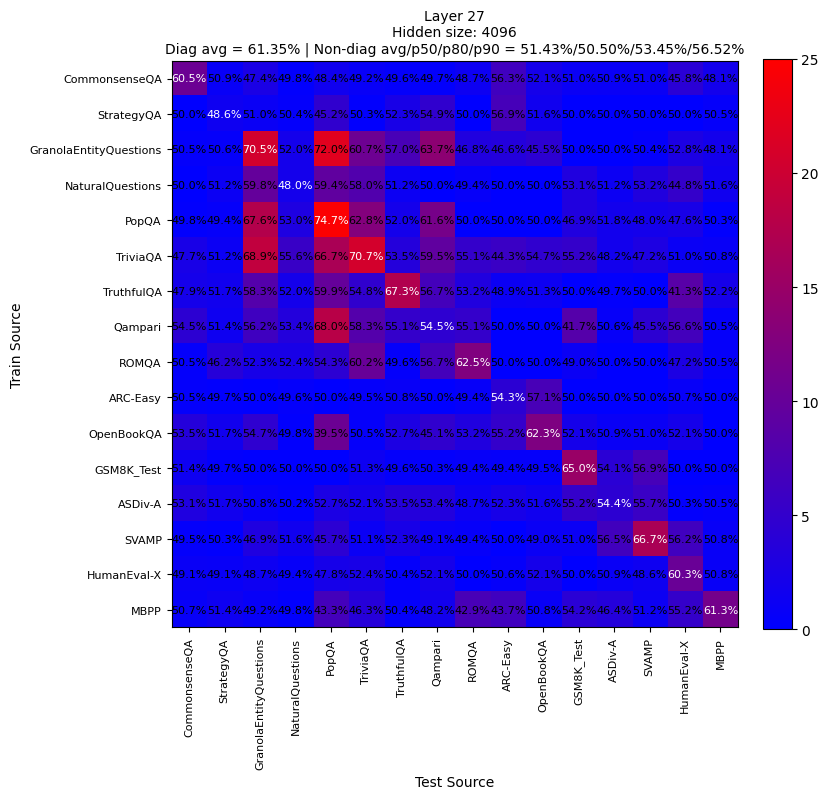} %
        \caption{Correctness prediction accuracy results of the classifier induced by $u_{27}(y-axis-dataset)$, using \mistralsevenb{}, while testing on the test set of the x-axis dataset.}
    \label{figure:cross_evaluation_mistral7_27}
    \end{minipage}
\end{figure}

\begin{figure}
    \centering
    \begin{minipage}{0.45\textwidth}
        \centering
        \includegraphics[width=1.1\textwidth]{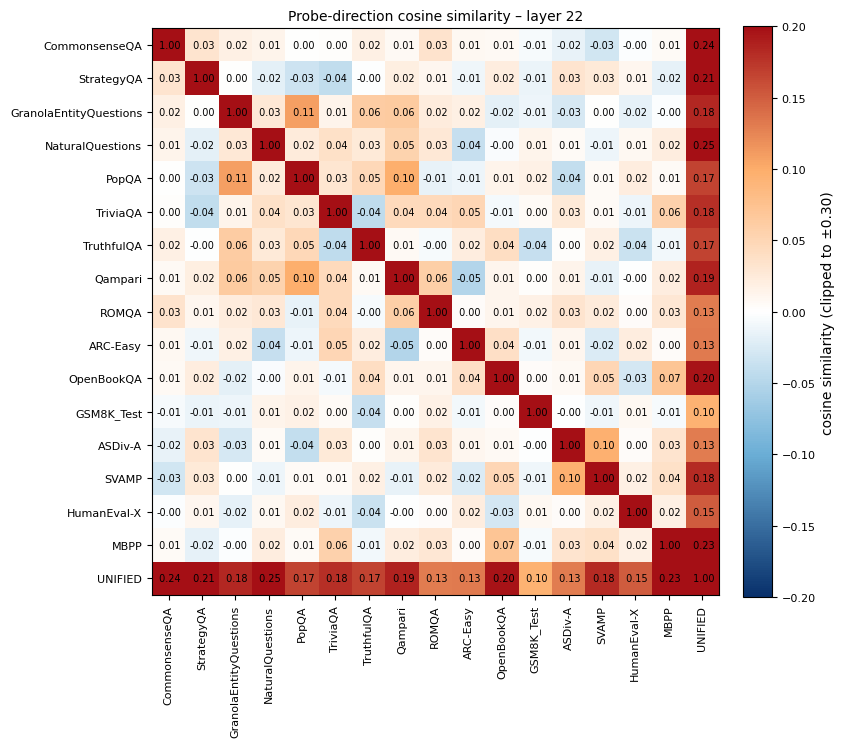} %
        \caption{Cosine similarity results across all linear uncertainty vectors at layer number 22 of \llamaeightb{}}
    \label{figure:cosine_similarities_llama8b_22}
    \end{minipage}\hfill
    \begin{minipage}{0.45\textwidth}
        \centering
        \includegraphics[width=1\textwidth]{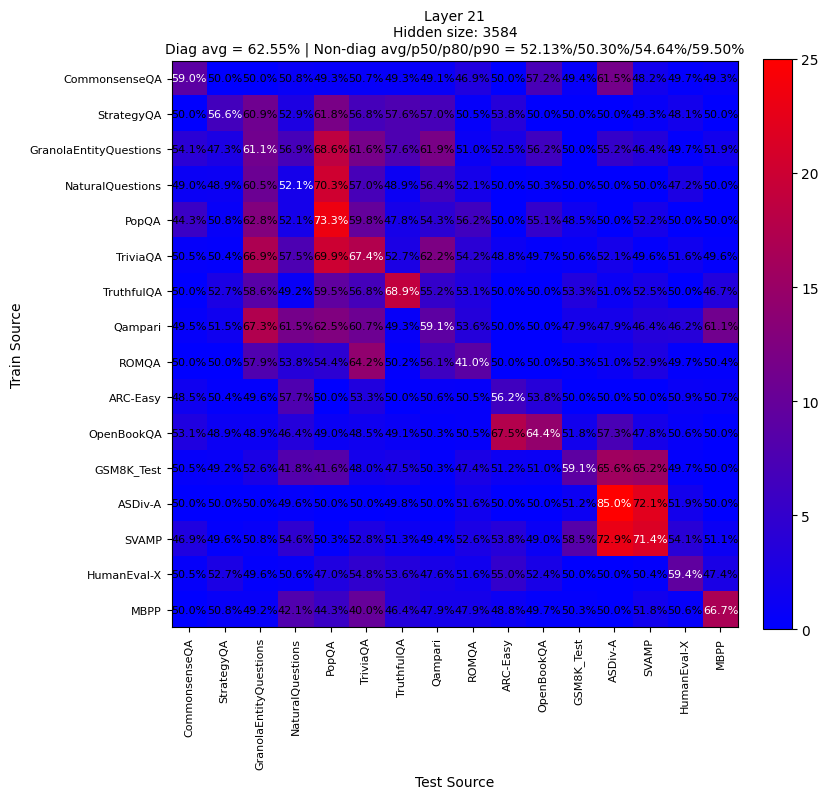} %
        \caption{Correctness prediction accuracy results of the classifier induced by $u_{21}(y-axis-dataset)$, using \qwensevenb{}, while testing on the test set of the x-axis dataset.}
    \label{figure:cross_evaluation_qwen7_21}
    \end{minipage}
\end{figure}

\section{LLMs Indeed Learn Different Types of Uncertainty}

In this section, we show that we can indeed find linear uncertainty vectors from which we can predict generation correctness to an extent that is better than random. We additionally claim and show that rather than learning one unified uncertainty, LLMs learn several  different ones. We later hypothesize that this fact might be one of the reasons for a high rate of misinformation and hallucinations that we observe generated by LLMs.

\subsection{The Concept of Uncertainty is Indeed Learned During Pretraining}

Table~\ref{table:acc_all_models_all_datasets_best_layer} presents the performance of our correctness classifiers, derived from the learned linear uncertainty vectors, across all evaluated models and datasets. While the uncertainty vector search is conducted independently at each transformer layer for every model–dataset pair, the table reports results from the best-performing layer only (a detailed layer-wise analysis is provided in a subsequent section). Notably, despite keeping the model weights entirely frozen and applying no further training, we are able to identify linear directions in the latent space that yield meaningful correctness predictions. The results demonstrate that, for a substantial number of datasets across all models, classification accuracy significantly exceeds the random baseline of 0.5. This provides strong empirical evidence that uncertainty is encoded within LLMs in a manner that is both learnable and linearly separable within their hidden representations.

\subsection{LLMs Learn Multiple Different Linear Uncertainty Vectors}

One of our key findings is that while linear uncertainty vectors can be identified across multiple layers in all examined models, these vectors are typically dataset-specific and distinct. Specifically, for a given layer $i$, a classifier induced from $\mathbf{u}_i(\mathcal{D}_1)$ often yields markedly different token-level correctness prediction accuracy across evaluation datasets compared to a classifier induced from $\mathbf{u}_i(\mathcal{D}_2)$, where $\mathcal{D}_1 \neq \mathcal{D}_2$. Furthermore, the cosine similarity between $\mathbf{u}_i(\mathcal{D}_1)$ and $\mathbf{u}_i(\mathcal{D}_2)$ is frequently near-zero, indicating near-linear independence between these vectors. Figure~\ref{figure:cross_evaluation_llama8_26} illustrates this effect for layer 26 of \llamaeightb{}, showing the accuracy of classifiers trained and tested on various datasets. Similarly, Figure~\ref{figure:cross_evaluation_mistral7_27} presents corresponding results for layer 27 of \mistralsevenb{}. In most cases, a vector trained on dataset $\mathcal{D}_1$ performs well when tested on $\mathcal{D}_1$, but approaches random performance when evaluated on other datasets. Additionally, Figure~\ref{figure:cosine_similarities_llama8b_22} shows cosine similarity scores among uncertainty vectors derived from \llamaeightb{} at layer 22. Aside from the unified classifier trained on a dataset union (\textsc{Unified}), nearly all vectors are close to orthogonal. In conjunction with the observation that most of the vectors can predict correctness substantially better than chance on at least one dataset, these results support the conclusion that LLMs encode uncertainty through multiple distinct and largely independent internal representations.

\subsection{Linear Uncertainty Topic Similarity}
An additional noteworthy finding emerges from an internal analysis of the submatrices corresponding to two sections of our dataset: \textbf{Fact-Lookup and Adversarial QA} and \textbf{Math Word Problems}. Specifically, when the induced classifier is evaluated on a dataset different from the one used to search for the uncertainty vector, it often attains a remarkably high accuracy—occasionally comparable to, or even surpassing, the level observed when the uncertainty vector is derived from the same dataset. This suggests that, for example, although mathematical uncertainty may be represented in various ways within the latent space, it is not strictly dataset-specific; rather, its semantic structure appears to be shared across tasks. This is further illustrated by the submatrix in Figure~\ref{figure:cross_evaluation_qwen7_21}, which includes three math benchmarks: GSM8K, ASDiv, and SVAMP. The results show that each uncertainty vector obtained from these datasets can substantially enhance the prediction of correctness across all three benchmarks when used as input for the induced classifier.

\subsection{Comparing to Zero-Shot Abstaining Skills}
As an additional evaluation of the uncertainty vectors, we assess their alignment with the model's own self-assessed knowledge through zero-shot prompting. Specifically, for each dataset question, we prompt the model to indicate whether it believes it knows the answer. We then measure the model's accuracy in this binary self-assessment and compute the Pearson correlation between these scores and the accuracy of our linear uncertainty-based correctness predictors. The resulting correlation coefficients are \textbf{0.45} for \llamaeightb{}, \textbf{0.38} for \mistralsevenb{}, and \textbf{0.42} for \qwenfortheenb{}. These findings indicate a substantial positive correlation, suggesting that the learned uncertainty vectors capture a meaningful signal related to the model's internal estimation of its own knowledge.

\section{In-Depth Analysis}
In this section, we analyze the gaps in performance of our induced correctness prediction classifiers, as a function of the layer number and the model size. We additionally study the effect of advanced training techniques such as instruction-tuning and [IDK]-tuning on the linear uncertainty encoding by the model. 

\subsection{Intermediate Layers are Usually More Exact}

\begin{figure}
    \centering
    \begin{minipage}{0.45\textwidth}
        \centering
        \includegraphics[width=0.7\textwidth]{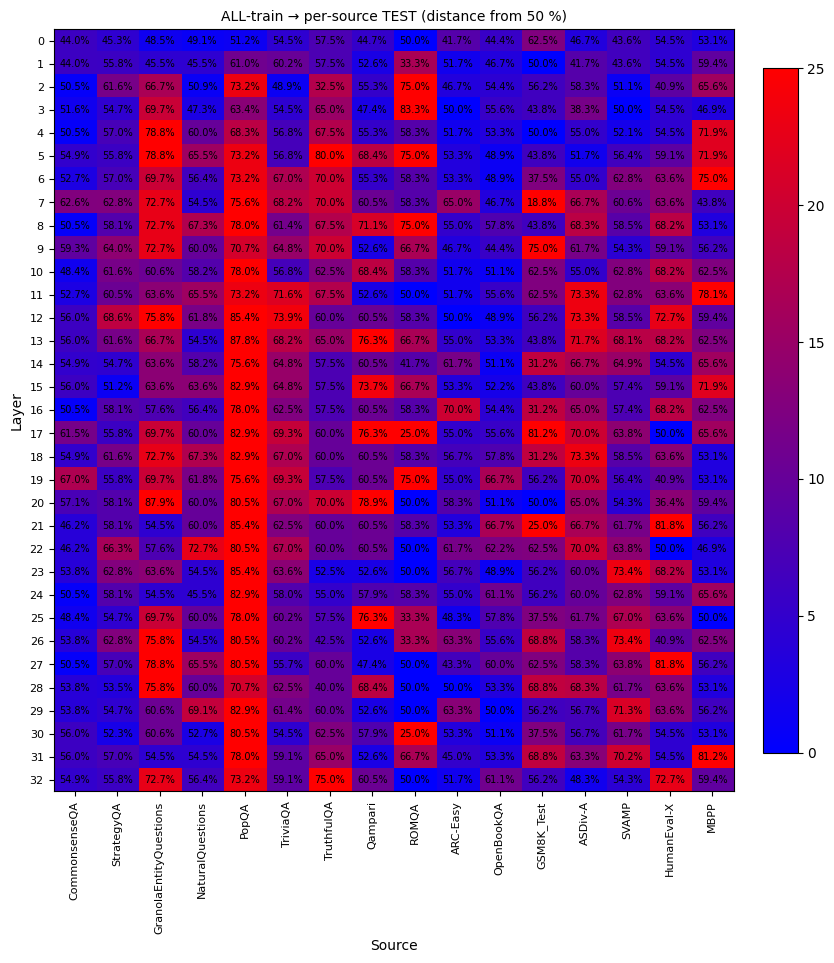} %
        \caption{Accuracy results of \mistralsevenb{} across all model layers and datasets. Here the induced classifiers were tested on the same dataset (but different split) as they were searched on.}
    \label{figure:acc_across_layers_datasets_mistral}
    \end{minipage}\hfill
    \begin{minipage}{0.45\textwidth}
        \centering
        \includegraphics[width=1\textwidth]{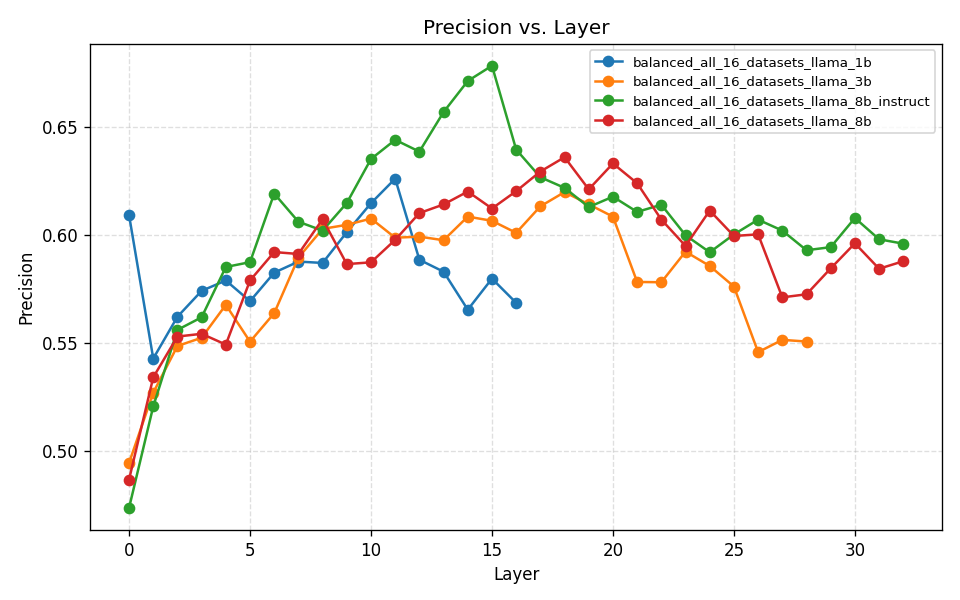} %
        \caption{Correctness prediction precision averaged over all datasets of the induced classifier, considering the Llama family: \llamaoneb{}, \llamathreeb{}, \llamaeightb{}, and \llamaeightbinstruct{}.}        
    \label{figure:llama_precision_layer_model_size_curve}
    \end{minipage}
\end{figure}

We begin by studying the behavior of uncertainty vectors and the corresponding correctness prediction performance across different transformer layers and model sizes. Figure~\ref{figure:acc_across_layers_datasets_mistral} reports the accuracy of uncertainty-based classifiers extracted from each layer of \mistralsevenb{}, evaluated on held-out splits of the same datasets used to induce them. On average, the vector from layer 17 achieves the highest prediction accuracy, with performance gradually declining in layers further from this point (noting that the model consists of 32 layers in total). This trend suggests that uncertainty-relevant information is most concentrated in intermediate layers. Complementing this, Figure~\ref{figure:llama_precision_layer_model_size_curve} shows the layer-wise average performance across multiple models, again highlighting that layers between $\frac{L}{2}$ and $\frac{3L}{4}$, where $L$ denotes the number of transformer layers, consistently yield the most reliable uncertainty signals. Notably, the precision results plotted in Figure~\ref{figure:llama_precision_layer_model_size_curve} show a marked drop in the final layers. This decline implies that the uncertainty vectors extracted from later layers tend to classify more incorrect generations as uncertain, indicating diminished model confidence in its own outputs.

\subsection{Size Doesn't Seem to Matter}

Figure~\ref{figure:llama_precision_layer_model_size_curve} illustrates the impact of model size on uncertainty-based correctness prediction accuracy across layers. Ignoring \llamaeightbinstruct{}, the highest performance is achieved by the classifier derived from layer 18 of \llamaeightb{}. Moreover, a comparison between \llamathreeb{} and \llamaeightb{} reveals negligible differences in average accuracy, suggesting comparable performance despite the disparity in model size. While \llamaoneb{} exhibits a slightly lower peak performance—approximately 1.1 points below the others—the overall trend indicates that the ability to represent uncertainty does not consistently improve with increasing model scale. These findings suggest that scaling alone is insufficient for enhancing uncertainty representation. In the subsequent section, we explore two training-based strategies that yield more substantial improvements.

\subsection{Boosting Uncertainty Capturing via Instruction-Tuning and [IDK]-tuning}

\begin{figure}
    \centering
    \begin{minipage}{0.45\textwidth}
        \centering
        \includegraphics[width=0.8\textwidth]{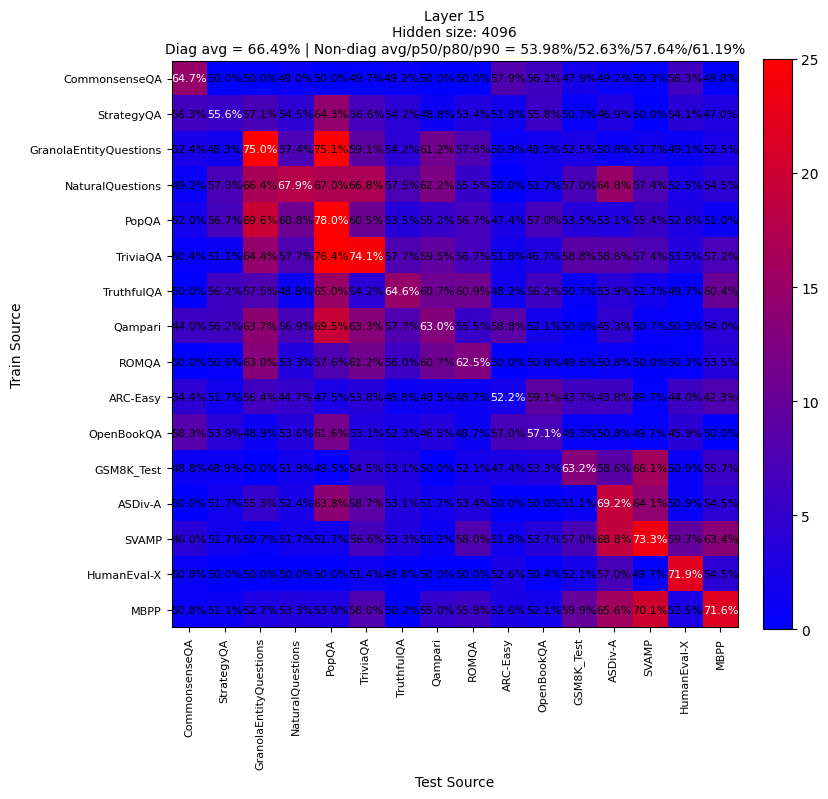} %
        \caption{Correctness prediction accuracy results of the classifier induced by $u_{15}(y-axis-dataset)$, using \llamaeightbinstruct{}, while testing on the test set of the x-axis dataset.}
    \label{figure:llama_instruct_acc_across_layers_datasets}
    \end{minipage}\hfill
    \begin{minipage}{0.45\textwidth}
        \centering
        \includegraphics[width=1\textwidth]{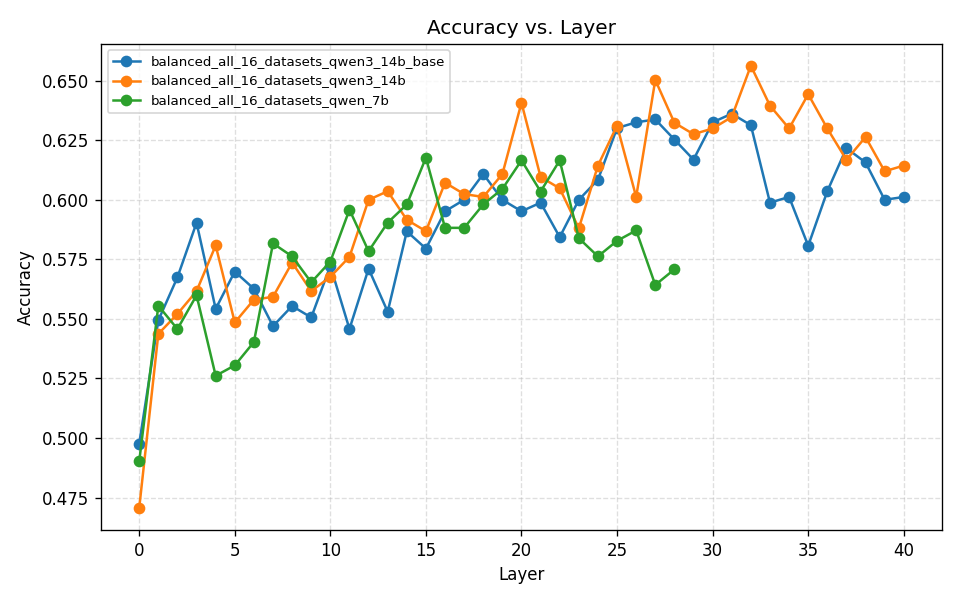} %
        \caption{Correctness prediction accuracy averaged over all datasets of the induced classifier, considering the Qwen family: \qwensevenb{}, \qwenfortheenb{}, and \qwenfortheenbinstruct{}}
    \label{figure:qwens_layer_model_size_curve_all}
    \end{minipage}
\end{figure}

\paragraph{Instruction-Tuning.}
Figure~\ref{figure:llama_precision_layer_model_size_curve} compares the performance of base LLaMA models—\llamaoneb{}, \llamathreeb{}, and \llamaeightb{}—with the instruction-tuned variant \llamaeightbinstruct{}, in terms of the correctness prediction accuracy derived from uncertainty vectors. The y-axis represents the average accuracy across all evaluated datasets. Notably, \llamaeightbinstruct{} consistently outperforms its base counterparts, indicating that instruction-tuning significantly enhances the model’s ability to encode and leverage uncertainty signals. A similar pattern is observed in Figure~\ref{figure:qwens_layer_model_size_curve_all}, where the instruction-tuned \qwenfortheenbinstruct{} demonstrates improved performance over the base \qwenfortheenb{}. Additionally, in both cases, the peak accuracy of the instruction-tuned models occurs several layers earlier than in their foundational equivalents, suggesting that instruction-tuning may facilitate earlier emergence of uncertainty-relevant representations within the model’s architecture.

\paragraph{[IDK]-Tuning.}

\begin{figure}
    \centering
    \begin{minipage}{0.45\textwidth}
        \centering
        \includegraphics[width=1\textwidth]{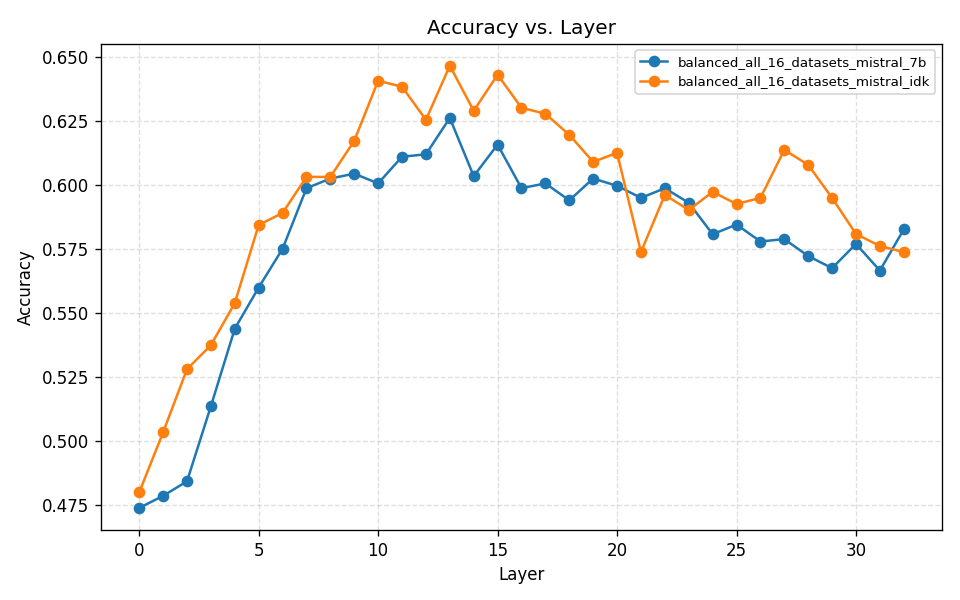} %
        \caption{Correctness prediction accuracy averaged over all datasets of the induced classifier, comparing \mistralsevenb{} against \idktunedmistral{}}
    \label{figure:mistral_layer_model_size_curve_all}
    \end{minipage}\hfill
    \begin{minipage}{0.45\textwidth}
        \centering
        \includegraphics[width=1\textwidth]{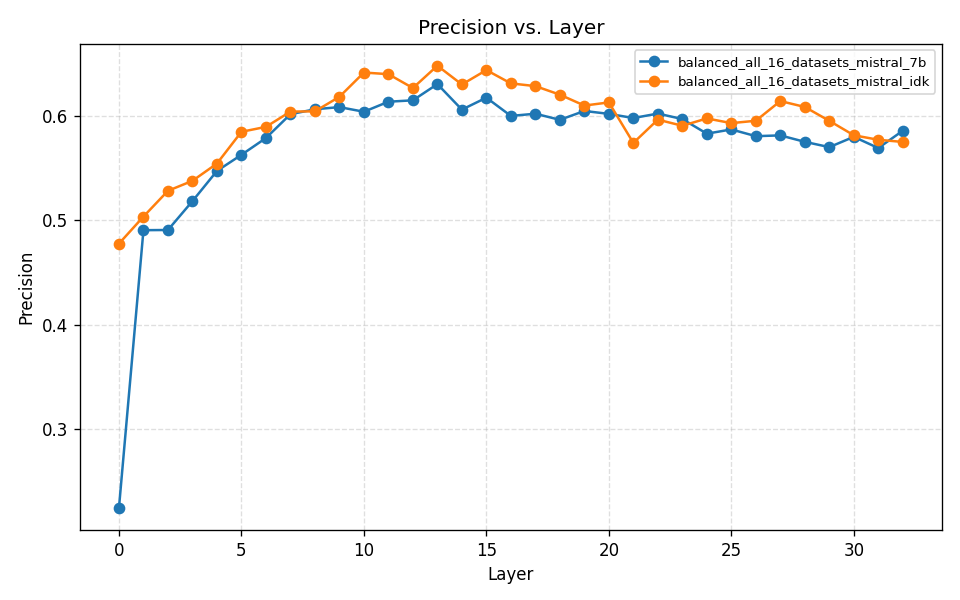} %
        \caption{Correctness prediction precision averaged over all datasets of the induced classifier, comparing \mistralsevenb{} against \idktunedmistral{}}
    \label{figure:mistral_layer_model_size_curve_precision}
    \end{minipage}
\end{figure}

Similar to instruction-tuning, [IDK]-tuning exerts notable influence on the model’s ability to capture uncertainty. This is reflected in the improved effectiveness of the resulting uncertainty vectors, which yield higher correctness prediction accuracy and reach peak performance in earlier layers of the model. These trends are illustrated in Figure~\ref{figure:mistral_layer_model_size_curve_all}. Additionally, precision scores shown in Figure~\ref{figure:mistral_layer_model_size_curve_precision} reveal a substantial gap at the first model layer. Specifically, the correctness predictors derived from the initial layer in the untuned model exhibit poor precision, indicating limited ability to detect generation errors and suggesting overconfidence at this early stage. [IDK]-tuning appears to mitigate this issue by aligning the initial layers more effectively with uncertainty signals. 

Additional phenomena we note is that for both these methods, we observe better cross-dataset results (that is, testing a vector that has been derived from dataset $D$, on a different dataset test split). Namely, each of the vectors has better generalization skills. This is shown in Figure~\ref{figure:llama_instruct_acc_across_layers_datasets}. 

\section{Related Work}

\paragraph{Model Calibration.}
\label{sec:related_calib}
Our analysis is closely related to the key challenge of model calibration~\citep{pmlr-v70-guo17a}: to provide a measure of the probability that a prediction is incorrect alongside the actual prediction. The problem of factual error detection can be viewed as a variation of calibration, where instead of a continuous probability, we provide a binary prediction for whether the model is correct or not. Common approaches to calibration are to perform various transformations on a model's output logits \citep{desai2020calibration, jiang-etal-2021-know} and measuring uncertainty \citep[e.g., see][]{kuhn2023semantic}. 
More recent works have studied the use of LMs for providing calibration, by training them on statements known to be factually correct or incorrect. This ``supervised'' approach has been explored via fine-tuning \citep{Kadavath2022LanguageM, lin2022teaching}, in-context learning \citep{cohen-etal-2023-crawling, alivanistos2022prompting}, zero-shot instruction-oriented \citep{cohen-etal-2023-lm} and consistency sampling \citep{yoran-etal-2023-answering} techniques.
Further recent studies \citep{azaria-mitchell-2023-internal} use the internal state of the model for classifying whether it is certain or not, use a new token for unanswerable inputs \citep{lu-etal-2022-controlling}, or construct a specific dataset for effectively tuning the model for answering refusal \citep{zhang-etal-2024-r}. Our work takes an analysis approach trying to better figure out the dynamics of the uncertainty encoding of pretrained models as well as better calibrated models. 

\paragraph{Mechanistic Interpretability} 

Recent work has been aiming to identify circuits and features within models that correspond to interpretable concepts such as factual recall, syntax, or positional reasoning \citep{olsson2022context, yu-etal-2023-characterizing}. For instance, tools such as SAE (Sparse Autoencoders) have been used to isolate human-interpretable features from residual stream activations \citep{meng2022locating}. Other studies explore how knowledge is stored and manipulated across layers, such as tracing factual associations or memorized content to specific directions in the latent space \citep{geva-etal-2021-transformer, gurnee2023finding, geva-etal-2023-dissecting, yu2024mechanistic}. Despite promising progress, full mechanistic understanding remains an open challenge due to the scale and complexity of modern models.

\section{Conclusion}

In this work, we present a framework for probing uncertainty representations within LLMs by identifying linear vectors in their latent space that predict generation correctness. Our findings establish that LLMs internalize uncertainty as a learnable and linearly accessible concept, one that can be extracted without fine-tuning the model weights. Moreover, we demonstrate that rather than encoding a singular notion of uncertainty, these models store multiple distinct uncertainty representations, each sensitive to the type of data and task. This multiplicity—often manifesting in nearly orthogonal vectors—suggests an underlying explanation for some of the inconsistencies and hallucinations commonly observed in LLM outputs.

Beyond the foundational discovery of uncertainty encoding, our analysis sheds light on the architectural and training factors that influence this phenomenon. We show that intermediate layers, regardless of model size, are the most predictive regions for uncertainty, and that larger models do not necessarily perform better at capturing it. More importantly, we find that instruction-tuning and [IDK]-tuning significantly enhance the model’s uncertainty awareness—both in accuracy and in early-layer alignment—pointing to training strategy, rather than scale, as the more critical lever for improving reliability. Our results offer actionable insights for both understanding and mitigating LLM hallucinations, and open up new directions for principled model design and interpretability.

\bibliographystyle{plainnat}
\bibliography{anthology,custom}

\appendix

\section{Limitations}
\label{sec:limitations}

While our analysis provides compelling evidence for the existence of linearly accessible uncertainty representations in LLMs, it is limited to linear probes and does not explore more complex, nonlinear structures that may further explain model behavior. Our evaluation focuses on a fixed set of models and datasets, which, although diverse, may not capture the full variability seen in real-world applications or domain-specific tasks. Additionally, correctness is treated as a proxy for uncertainty, which may not fully align with how uncertainty manifests in open-ended or ambiguous generation scenarios. Finally, the performance of our classifiers may also be influenced by dataset-specific biases, potentially limiting generalizability.

\section{Computer Resources}
\label{sec:comp_resources}

In our experiments we use one NVIDIA A100 80G GPU. 

\newpage
\section*{NeurIPS Paper Checklist}

The checklist is designed to encourage best practices for responsible machine learning research, addressing issues of reproducibility, transparency, research ethics, and societal impact. Do not remove the checklist: {\bf The papers not including the checklist will be desk rejected.} The checklist should follow the references and follow the (optional) supplemental material.  The checklist does NOT count towards the page
limit. 

Please read the checklist guidelines carefully for information on how to answer these questions. For each question in the checklist:
\begin{itemize}
    \item You should answer \answerYes{}, \answerNo{}, or \answerNA{}.
    \item \answerNA{} means either that the question is Not Applicable for that particular paper or the relevant information is Not Available.
    \item Please provide a short (1–2 sentence) justification right after your answer (even for NA). 
\end{itemize}

{\bf The checklist answers are an integral part of your paper submission.} They are visible to the reviewers, area chairs, senior area chairs, and ethics reviewers. You will be asked to also include it (after eventual revisions) with the final version of your paper, and its final version will be published with the paper.

The reviewers of your paper will be asked to use the checklist as one of the factors in their evaluation. While "\answerYes{}" is generally preferable to "\answerNo{}", it is perfectly acceptable to answer "\answerNo{}" provided a proper justification is given (e.g., "error bars are not reported because it would be too computationally expensive" or "we were unable to find the license for the dataset we used"). In general, answering "\answerNo{}" or "\answerNA{}" is not grounds for rejection. While the questions are phrased in a binary way, we acknowledge that the true answer is often more nuanced, so please just use your best judgment and write a justification to elaborate. All supporting evidence can appear either in the main paper or the supplemental material, provided in appendix. If you answer \answerYes{} to a question, in the justification please point to the section(s) where related material for the question can be found.

IMPORTANT, please:
\begin{itemize}
    \item {\bf Delete this instruction block, but keep the section heading ``NeurIPS Paper Checklist"},
    \item  {\bf Keep the checklist subsection headings, questions/answers and guidelines below.}
    \item {\bf Do not modify the questions and only use the provided macros for your answers}.
\end{itemize}


\begin{enumerate}

\item {\bf Claims}
    \item[] Question: Do the main claims made in the abstract and introduction accurately reflect the paper's contributions and scope?
    \item[] Answer: \answerYes{} 
    \item[] Justification: We run extensive experiments to support our claims.

\item {\bf Limitations}
    \item[] Question: Does the paper discuss the limitations of the work performed by the authors?
    \item[] Answer: \answerYes{} 
    \item[] Justification: See \autoref{sec:limitations}.

\item {\bf Theory assumptions and proofs}
    \item[] Question: For each theoretical result, does the paper provide the full set of assumptions and a complete (and correct) proof?
    \item[] Answer: \answerNA{} 
    \item[] Justification: N/A

    \item {\bf Experimental result reproducibility}
    \item[] Question: Does the paper fully disclose all the information needed to reproduce the main experimental results of the paper to the extent that it affects the main claims and/or conclusions of the paper (regardless of whether the code and data are provided or not)?
    \item[] Answer: \answerYes{} 
    \item[] Justification: See section~\ref{sec:setup} and~\ref{sec:expermental_setup}.

\item {\bf Open access to data and code}
    \item[] Question: Does the paper provide open access to the data and code, with sufficient instructions to faithfully reproduce the main experimental results, as described in supplemental material?
    \item[] Answer: \answerNo{} 
    \item[] Justification: Will be released with the camera-ready.

\item {\bf Experimental setting/details}
    \item[] Question: Does the paper specify all the training and test details (e.g., data splits, hyperparameters, how they were chosen, type of optimizer, etc.) necessary to understand the results?
    \item[] Answer: \answerYes{}{} 
    \item[] Justification: See section~\ref{sec:datasets}.

\item {\bf Experiment statistical significance}
    \item[] Question: Does the paper report error bars suitably and correctly defined or other appropriate information about the statistical significance of the experiments?
    \item[] Answer: \answerNo{} 
    \item[] Justification: Our experiments have been conducted at large scale including 9 different models and 16 different datasets, means that it's very likely that statistical errors are negligible in this case. 

\item {\bf Experiments compute resources}
    \item[] Question: For each experiment, does the paper provide sufficient information on the computer resources (type of compute workers, memory, time of execution) needed to reproduce the experiments?
    \item[] Answer: \answerYes{} 
    \item[] Justification: See \autoref{sec:comp_resources}
    
\item {\bf Code of ethics}
    \item[] Question: Does the research conducted in the paper conform, in every respect, with the NeurIPS Code of Ethics \url{https://neurips.cc/public/EthicsGuidelines}?
    \item[] Answer: \answerYes{} 
    \item[] Justification: We reviewed the guidelines and are in conformance.

\item {\bf Broader impacts}
    \item[] Question: Does the paper discuss both potential positive societal impacts and negative societal impacts of the work performed?
    \item[] Answer: \answerYes{} 
    \item[] Justification: See \autoref{sec:limitations}.
    
\item {\bf Safeguards}
    \item[] Question: Does the paper describe safeguards that have been put in place for responsible release of data or models that have a high risk for misuse (e.g., pretrained language models, image generators, or scraped datasets)?
    \item[] Answer: \answerNo{} 
    \item[] Justification: We do not include safeguards.

\item {\bf Licenses for existing assets}
    \item[] Question: Are the creators or original owners of assets (e.g., code, data, models), used in the paper, properly credited and are the license and terms of use explicitly mentioned and properly respected?
    \item[] Answer: \answerYes{} 
    \item[] Justification: We cite all used models, datasets and methods. 

\item {\bf New assets}
    \item[] Question: Are new assets introduced in the paper well documented and is the documentation provided alongside the assets?
    \item[] Answer: \answerYes{} 
    \item[] Justification: Full details of all assets will be made available.

\item {\bf Crowdsourcing and research with human subjects}
    \item[] Question: For crowdsourcing experiments and research with human subjects, does the paper include the full text of instructions given to participants and screenshots, if applicable, as well as details about compensation (if any)? 
    \item[] Answer: \answerNA{}{} 
    \item[] Justification: N/A

\item {\bf Institutional review board (IRB) approvals or equivalent for research with human subjects}
    \item[] Question: Does the paper describe potential risks incurred by study participants, whether such risks were disclosed to the subjects, and whether Institutional Review Board (IRB) approvals (or an equivalent approval/review based on the requirements of your country or institution) were obtained?
    \item[] Answer: \answerNA{} 
    \item[] Justification: N/A

\item {\bf Declaration of LLM usage}
    \item[] Question: Does the paper describe the usage of LLMs if it is an important, original, or non-standard component of the core methods in this research? Note that if the LLM is used only for writing, editing, or formatting purposes and does not impact the core methodology, scientific rigorousness, or originality of the research, declaration is not required.
    \item[] Answer: \answerYes{} 
    \item[] Justification: See \autoref{sec:expermental_setup}.

\end{enumerate}

\end{document}